\documentclass[runningheads]{llncs}
\usepackage[T1]{fontenc}
\usepackage{lmodern}
\usepackage{graphicx}
\usepackage{url}
\usepackage{cite}
\usepackage{tabularx}
\usepackage{amsmath}
\usepackage{wrapfig}
\usepackage{amssymb}
\usepackage{algorithm2e}
\SetKwComment{Comment}{/* }{ */}
\RestyleAlgo{ruled} 
\usepackage{bbm}
\usepackage{wrapfig}
\usepackage[misc,geometry]{ifsym}
\usepackage{multirow,tabularx}
\usepackage{tabularx,lipsum}
\usepackage{booktabs}
\usepackage{enumitem}
\usepackage{svg}

\usepackage [autostyle, english = american]{csquotes}
\usepackage [USenglish]{babel}

\usepackage{microtype}

\usepackage[pdftex, colorlinks=true, hyperfootnotes=true, hyperindex=true,
plainpages=false, pagebackref=false, pdfpagelabels=true, pdfstartview=FitH,
linkcolor=blue, citecolor=blue, urlcolor=blue,
bookmarks, bookmarksopen, bookmarksdepth=3]{hyperref}
\usepackage{orcidlink}
\usepackage{multirow}

\makeatletter
\DeclareRobustCommand{\textsupsub}[2]{{%
  \m@th\ensuremath{%
    ^{\mbox{\fontsize\sf@size\z@#1}}%
    _{\mbox{\fontsize\sf@size\z@#2}}%
  }%
}}
\makeatother

\counterwithout{table}{chapter}
\begin{document}
\definecolor{orcidlogocol}{HTML}{A6CE39}
\title{Attention Please: What Transformer Models Really Learn for Process Prediction}
\titlerunning{What Transformer Models Really Learn for Process Prediction}
%
\author{Martin Käppel\inst{1}\orcidlink{0009-0003-3420-8037} \small{\Letter} \and
Lars Ackermann\inst{1}\orcidlink{0000-0002-6785-8998} \and
Stefan Jablonski\inst{1} \and Simon Härtl\inst{1}}

\authorrunning{M. Käppel et al.}
%
\institute{Institute for Computer Science, University of Bayreuth, Bayreuth, Germany \\
	\email{\{martin.kaeppel, lars.ackermann, stefan.jablonski, simon.haertl\}@uni-bayreuth.de}\\
}
\maketitle              
\vspace{-20pt}
\begin{abstract}
Predictive process monitoring aims to support the execution of a process during runtime with various predictions about the further evolution of a process instance. In the last years a plethora of deep learning architectures have been established as state-of-the-art for different prediction targets, among others the transformer architecture. The transformer architecture is equipped with a powerful attention mechanism, assigning attention scores to each input part that allows to prioritize most relevant information leading to more accurate and contextual output. However, deep learning models largely represent a black box, i.e., their reasoning or decision-making process cannot be understood in detail. This paper examines whether the attention scores of a transformer based next-activity prediction model can serve as an explanation for its decision-making. We find that attention scores in next-activity prediction models can serve as explainers and exploit this fact in two proposed graph-based explanation approaches. The gained insights could inspire future work on the improvement of predictive business process models as well as enabling a neural network based mining of process models from event logs.

\end{abstract}

\section{Introduction}
\label{sec:introduction}
\emph{Predictive Business Process Monitoring (PBPM)} provides runtime support for the
execution of a process 
with various predictions about the further evolution of a process instance, based on predictive models created from event logs~\cite{DiFrancescomarino2022}. Early knowledge of the likely course of a process instance offers significant advantages: upcoming steps can be prepared, and potential problems can be identified and mitigated early, e.g., in resource and time planning~\cite{DiFrancescomarino2022}. In the last years a plethora of deep learning architectures have been established as state-of-the-art for different prediction targets (e.g., next activity, outcome, remaining time), among others Convolutional Neural Networks~\cite{Pasquadibisceglie2019}, Long Short Term Memory Neural Networks (LSTM)~\cite{Evermann2017,Camargo2019}, or more recently transformer architectures~\cite{Bukhsh2021}. 

However, deep learning models largely represent a black box, i.e., their reasoning or decision-making process is hidden~\cite{Nauta2023}. Consequently, users often do not trust them, which hinders their application~\cite{Carvalho2019}. To address this issue, the discipline of \emph{explainable artificial intelligence (XAI)} tries to find explanations for the decisions of machine learning models~\cite{Carvalho2019,Nauta2023}. Such approaches also emerged in the context of PBPM~\cite{Weinzierl2020}. Depending on whether an explainer aims to explain the prediction for a single running process instance or a collection of instances (e.g., an event log) we distinguish between \emph{local} and \emph{global explainers}. While global approaches investigate, how a model makes decisions in general, local approaches examine only individual predictions for a particular input. Previous research has mainly focused on local explainers, so that it still remains unclear, whether deep learning models implicitly learn the underlying process from the event log. The landmark work of Evermann et. al.~\cite{Evermann2017} proposes the hypothesis that deep learning models for next activity prediction implicitly learn the process structure. Peeperkorn et. al. narrowed this research gap by investigating this hypothesis for LSTMs and revealed that they \textit{"can struggle to learn process model structure"}~\cite{Peeperkorn2023}. Nevertheless, the investigation of this hypothesis for other architectures seems to be worthwhile~\cite{Peeperkorn2023}, especially since LSTMs have issues with long sequences, as they pay elements that lie far behind less attention. 
The transformer architecture~\cite{Vaswani2017} bypasses this issue using an attention mechanism, assigning attention scores to each input part and prioritizing those with the highest scores as the most relevant, regardless of their position~\cite{Bukhsh2021,Vaswani2017}. The transformer architecture has spurred new state-of-the-art approaches across various research domains~\cite{Vaswani2017} (e.g., language models such as GPT-3 and BERT). 

As this architecture also excels in PBPM to predict the control-flow in form of the next activity~\cite{Bukhsh2021}, this paper explores 
the following research question: \textit{Do the attention scores of the transformer architecture give any insights whether the trained prediction model achieved an understanding of the control flow of a process?}
Thus, our main contributions are: \textbf{(I)} We examine whether the aforementioned attention scores of the transformer architecture can serve as a solid basis for the development of XAI approaches. \textbf{(II)} We contribute two global, transformer-specific explanation approaches to investigate the explainability of the transformer's attention mechanism and to offer types of explanation, which are motivated rather by the BPM domain than the AI domain. \textbf{(III)} Using eight data sets and five complementary metrics we evaluate the quality of the explanatory approaches, thereby offering a benchmark for sustainable comparability. 

The rest of the paper is structured as follows: Section~\ref{sec:preliminaries} provides background on PBPM (i.e., basic terminology, fundamentals of the transformer architecture). After delimiting our approach from related work (Section \ref{sec:related-work}), a preliminary study investigates, whether attention scores can serve as an explanation (Section~\ref{sec:pre-study}). 
Based on the result, we introduce two global explanation approaches (Section~\ref{sec:concept}), systematically applied to various prediction models trained on real-life event logs (Section~\ref{sec:evaluation}). Section~\ref{sec:conclusion-and-future-work} discusses potential limitations and concludes the paper.

\section{Preliminaries}
\label{sec:preliminaries}


\subsection{Basic Terminology}
\label{subsec:basic-terminology}
In the following $A$ denotes the set of viable activities in a business process. The execution of a (business) process can be recorded in form of a \emph{trace} $\sigma = \langle e_1, ..., e_n \rangle$, i.e., a temporally ordered sequence of events associated with the same process instance. Each \emph{event} represents the execution of an activity in the process and is characterized by event attributes, such as the name of the corresponding activity. Without loss of generality (since the considered process transformer only process activities), we denote traces in the following as sequences of their activity names. We denote with $\pi_A$ a function that returns for a given event its executed activity and with $\vert\sigma\vert$ the length of the trace $\sigma$. The set of all traces pertaining to the same (business) process is referred as \emph{(process) event log}. For PBPM approaches, prefixes of a trace are used to represent running process instances:
\begin{definition}[\textbf{Prefix}]
Let $\sigma = \langle e_1, ..., e_n \rangle$ denote a trace and let be $r \in \{1, ..., n-1\}$. The \textbf{prefix} of length $r$ is a function $hd$ which returns the first $r$ elements of a trace: $hd(\sigma, r) = \langle e_1, ..., e_r \rangle$. 
\end{definition}
This prefix, is then used as input to a function $\Omega$ predicting the next activity, given by $\Omega(hd(\sigma, r)) = \pi_A(e_{r+1})$. A PBPM approach, in this paper a transformer model, aims to learn the function $\Omega$ using a given event log.

\subsection{Attention Mechanism}\label{subsec:transformer-architecture}
In deep learning the attention mechanism allows models to dynamically focus on different parts of a input sequence for generating output. The fundamental idea behind attention is to learn to assign different scores to the parts of the input, indicating how much focus the model should pay them when performing a task. Hence, it allows to prioritize most relevant information leading to more accurate and contextual output. 
The approach investigated in this paper is based on the so called \emph{self-attention mechanism}~\cite{Vaswani2017} utilized within the transformer architecture for next activity prediction proposed by~\cite{Bukhsh2021}. In the following, we describe parts of the architecture relevant for the paper at hand, i.e., the input, the self-attention mechanism, and the output of the transformer. For a description of the remaining parts of the architecture, we refer to~\cite{Vaswani2017} and~\cite{Bukhsh2021}, respectively.

According to~\cite{Bukhsh2021} the transformer receives as input a list of activities (i.e., a prefix). This input is then embedded into a high-dimensional space of dimension $d_k$ by converting each element of the list into a so called \emph{embedding vector}. This representation leads to a continuous representation of the input and captures positional, semantic, and syntactic properties of the corresponding elements, and serves as input for the self-attention. It is important to note that the embeddings are automatically learned during training. In the attention mechanism an embedding vector $x_i$ is used in three different ways: as \emph{query}, \emph{key}, and \emph{value vector}. We get them by multiplying $x_i$ with matrices $W_Q$, $W_K$, and $W_V$ that are learned during the training process, and pack them for more efficiency into matrices $Q$, $V$, and $K$ (each column represents one vector)~\cite{Vaswani2017}. Next, we use $Q$, $K$, and $V$ to calculate an attention score reflecting the importance of each element (query) under consideration in relation to the others. First, each element of the input sequence is scored against the currently considered element by taking the matrix product $QK^T$. The values of the resulting matrix are divided by $\sqrt{d_k}$ before applying a row-wise softmax operation to normalize the values to avoid slow training through vanishing gradients~\cite{Vaswani2017}. The obtained matrix after the softmax operation is called \emph{attention score matrix} or \emph{attention matrix} for short. 
\begin{wrapfigure}{r}{0.5\textwidth}
    \centering

    \includegraphics[scale=0.5]{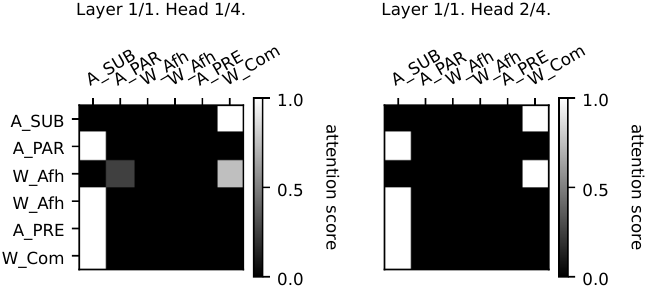}

    \caption{Attention scores of two different heads for a prefix of the BPIC12 event log.}
  
    \label{fig:sample-attention-score-bpic12}
\end{wrapfigure}
Finally each value vector is multiplied by the attention scores. The intent behind this step is, that if the attention value is very small, it eliminates the value vector due to irrelevance. Hence, the calculation can be summarized as:

\begin{flalign*}
    &Att(Q, K, V) = softmax \left(\frac{QK^T}{\sqrt{d_k}} \right) V.&&
\end{flalign*}

To learn different types of relationships and patterns between the elements of the input sequence, we use $h$ independent and parallel attention mechanisms (called \emph{attention heads} or \emph{heads}) and combine them to so called \emph{multi-head attention}~\cite{Vaswani2017}: $MultiHeadAttention(Q, K, V) = concat(h_1, ... h_h)W^O$
with $h_j$ = $Att\left(QW_Q^{j}, KW_K^{j}, VW_V^{j}\right)$. In this formula the resulting matrix of each head $j$ is arranged side by side and multiplied by a weight matrix $W^O$, that is learned during training. 
The process transformer presented in~\cite{Bukhsh2021} uses four heads. The results of the multi-head attention are further processed and aggregated with different layers ending in a softmax output layer $O$~\cite{Bukhsh2021}. This layer has a neuron for each activity in the process and accordingly returns a probability distribution $p$ over the activities (also known as \emph{prediction vector}). The activity with the highest probability is extracted as predicted next activity using the argmax function. 

In the following let $\mathcal{M}$ be the trained transformer model. We denote with $p_{\sigma} = \mathcal{M}(\sigma)$ the models' outputted probability distribution for a prefix $\sigma$. Moreover, $p_{\sigma}(a)$ refers to the prediction score for activity $a \in \mathcal{A}$ in $p_{\sigma}$. With $M_{\sigma}^{j}$ we denote the attention score matrix of head $h_j$ received for prefix $\sigma$. To access a particular attention score within $M_{\sigma}^{j}$, we write  $M_{\sigma}^{j}(i,k$) to access the score in the $i$-th row and the $k$-th column. The vector $M_{\sigma} = (M_{\sigma}^{1}, ..., M_{\sigma}^{h})$ summarizes the attention score matrices of all heads in $\mathcal{M}$. For further investigations, it is necessary to understand the attention scores for the elements of the input sequence as a probability distribution. Therefore we flatten the attention score matrix $M_{\sigma}^{j}$ for head $j$ into a vector $v_{\sigma}^{j} = (M_{\sigma}^{1}(1,1), M_{\sigma}^{1}(1,2), ..., M_{\sigma}^{1}(1,\vert\sigma\vert), M_{\sigma}^{2}(1,\vert\sigma\vert), ..., M_{\sigma}^{h}(\vert\sigma\vert,\vert\sigma\vert)$ (concatenating the rows of $M_{\sigma}^{j}$). The flattened representation across all heads is then defined as $v_{\sigma} = (v_{\sigma}^{1}, ...,v_{\sigma}^{h})$. Based on the flattened representations we define the \emph{head attention score distribution} for head $j$ obtained from $\sigma$ as $\alpha_{\sigma}^{j} = \frac{v_{\sigma}^{j}}{\|v_{\sigma}^{j}\|}$, and the \emph{attention score distribution} across all heads as $\alpha_{\sigma} = \frac{v_{\sigma}}{\|v_{\sigma}\|}$, where $\|\cdot\|$ denotes the L1-norm, i.e., the component-wise sum of the elements.

\subsection{Interpretation of Attention Scores}
Fig.~\ref{fig:sample-attention-score-bpic12} depicts attention score matrices for two heads, illustrating how elements of an input sequence (here the activities of the events) relate to each other through the model's attention mechanism. The interpretation of an attention score matrix is identically for all heads. In the attention score matrix, rows and columns correspond to elements of the input sequence, with rows representing these elements as queries and columns as keys. By examining a row, we can assess the importance the transformer model assigns to each other element in the input sequence from the perspective of the row's element. In contrast, analyzing a column reveals the degree of attention all other elements in the input sequence pay to the element associated with that column. Within the matrix higher attention scores appears lighter, while lower scores are shown darker, indicating the strength of the relationship between the elements in the respective row and column.






\vspace{-10pt}
\section{Related Work}
\label{sec:related-work}
In the following we briefly discuss XAI approaches in the PBPM domain with focus on next activity prediction. They can roughly be classified into \emph{model-specific} or \emph{model-agnostic} approaches. While model-specific approaches are tailored to the inner working of a model, model-agnostic approaches can be applied to any machine learning model, regardless of its internals. Although model-agnostic approaches are highly versatile and universally applicable, they provide less detailed insights in the decision processes of the particular model. The survey~\cite{Stierle2021} identifies 20 XAI approaches in the PBPM domain and categorizes them along different dimensions. It shows, that the vast majority of approaches is model-agnostic and limited to local explanations. Hence, there is a lack of global model-specific explainer. When combining criteria of model-specific and next-activity explainers, only three papers remain~\cite{Weinzierl2020,Sindhgatta2020,Hanga2020}, that can be considered as directly related work. None of these used the transformer as their specific model, however~\cite{Sindhgatta2020} uses an attention mechanism within an LSTM, whereas this mechanism differs from those used by the transformer from~\cite{Bukhsh2021}. The results in this approach show accumulated attention scores for each position in a prefix. Although the authors refer to the work of~\cite{Serrano2019} which states that attention cannot be interpreted ``as is``, they do not further consider tests proposed in~\cite{Serrano2019} to verify and critically reflect whether the observed attention scores are truly interpretable. 
Hence, it is still an open question in the PBPM domain, whether attention scores can be trusted as basis for explanation~\cite{Wiegreffe2019,Jain2019}. We address this question in Section~\ref{sec:pre-study}. The \emph{Explainable Next Activity Prediction (XNAP)} approach proposed in~\cite{Weinzierl2020} provides explanations for LSTMs for next activity prediction using layer-wise relevance propagation resulting in assigning each activity in the input prefix a relevance score. However, just as in~\cite{Sindhgatta2020} they do not conduct any quantitative evaluation and limit themselves to a qualitative analysis of the results. The approach proposed in~\cite{Hanga2020} also relies on LSTM networks but differs from~\cite{Sindhgatta2020} and~\cite{Weinzierl2020} by outputting a directed and weighted graph as explanation instead of relevance score for elements in the input prefix. In the resulting graph, activities are represented by nodes and the probabilities on each edge $(u,v)$ indicate the likelihood of the transition from $u$ to $v$ as predicted by the model. Hence, this graph can be considered only as a Directly Follows Graph enriched with probabilities. While this approach reflects which edges can be traversed according to the LSTM model, it is questionable whether the graph can explain the LSTM’s decision-making. Beside this direct related work there are some further noteworthy approaches to be shortly discussed as it serves as inspiration for the developed approaches within this paper. The LORE~\cite{Guidotti2018} approach approximates a black box model locally by generating local instances around the input instance. In this generation process the instances are mutated, i.e., changing continuous or categorical values. The LORELEY~\cite{Huang2022} approach extends the LORE approach~\cite{Guidotti2018} and applies it to next-activity prediction in the PBPM domain by restricting mutation options to avoid violating the control flow. However, it is a model-agnostic approach and therefore not comparable to the approach presented in this paper. 

Moreover, we distinguish our work from distantly related approaches that propose local methods or focus on other prediction targets. 
For example, Rizzi et. al.~\cite{Rizzi2020} identify features responsible for wrong outcome-predictions using the model-agnostic local explainers LIME and SHAP. By reducing the impact of the identified features, the accuracy of the prediction model can be increased. More general guidelines for the use of features are determined in the work of Stevens et. al~\cite{Stevens2022} that reveals for outcome-prediction that a small number of features that are as much as possible independent are preferable without compromising on faithfulness. In addition, Elkawaga et. al.~\cite{Elkhawaga2022} emphasise the importance of investigating the extent to which explanations reflect the characteristics of the data. Their findings indicate that feature selection is crucial for both improving prediction quality and obtaining reliable explanations. However, since the transformer investigated in our work, exclusively utilises the activity event attribute, feature selection and feature importance do not apply for creating explanations. This is one of the main reasons why a tailored XAI approach for this architecture is required. 
While the aforementioned approaches primarily rely on input perturbations,~\cite{Mehdiyev_2021} proposes a local explanation approach for outcome prediction that defines local regions in the latent space of the deep learning model via clustering and decision trees.

\section{Pre-Study: The Relevance of Attention Scores}
\label{sec:pre-study}

Before using attention scores as pivotal component in an explanation approach for next activity prediction, we verify their reliability in this context, i.e., that they indeed highlight activities decisive for a prediction. We investigate whether attention scores in general affect the prediction (Exp. \hyperref[experiment1]{1}) and shed light on the extent to which particular attention scores contribute to a prediction (Exp. \hyperref[experiment2]{2}).


\paragraph{\textbf{Datasets:}} For our study we use eight frequently used real-world event logs (see Table~\ref{used-event-logs}) obtained from the \emph{4TU Center for Research Data}\footnote{\url{https://data.4tu.nl/}}. The BPIC12 datasets (O, W, and WC) are obtained from the BPIC12 event log by different filtering operations. While BPIC12\_O is derived from BPIC12 by only keeping events whose activities start with "O\_", BPIC12\_W only contains events whose activities start with prefix "W\_". The BPIC12\_WC is a refinement of BPIC12\_W only keeping events whose lifecycle attribute is set to "complete".  


\begin{table}[t]
	\begin{tabularx}{\textwidth}{lXXXXXX}
		\toprule
		\small Event Log &  \#Cases & \#Act. & \#Events & AVG. $\vert\sigma\vert$ &  Max. $\vert\sigma\vert$ &   \#Variants    \\ \midrule
  BPIC12 & 13087 & 24 & 262200 & 20.04 & 175 & 4366 \\
  BPIC12\_O & 5015 & 7 & 31244 & 6.23 & 30 & 168 \\
  BPIC12\_W & 9658 & 7 & 170107 & 17.61 & 156 & 2643 \\
  BPIC12\_WC & 9658 & 6 & 72413 & 7.50 & 74 & 2263 \\
  BPIC13\_CP & 1487 & 4 & 6660 & 4.48 & 35 & 183 \\
  BPIC13-I & 7554 & 4 & 65533 & 8.68 & 123 & 1511 \\
  Helpdesk & 4580 & 14 & 21348 & 4.66 & 15 & 226 \\
  Sepsis & 1050 & 16 & 15214 & 14.49 & 185 & 846 \\
  \bottomrule
	\end{tabularx}

	\caption{Statistics of the event logs ($\vert\sigma\vert$ is the trace length). We focused only on control-flow relevant characteristics, since the transformer model only considers activities.}

	\label{used-event-logs}
\end{table}

\paragraph{\textbf{General experiment setup: }}
All sub-experiments in this study share some similarities in the setup. We divide each event log randomly into training and test data in a ratio of 70\% training to 30\% testing data. To ensure reproducibility of our experiments, seeds are set. To create specific test samples, we extract all prefixes with a minimum length of 1 from the traces in the test data. The transformer model is trained on the training data using the standard hyperparameters as defined in~\cite{Bukhsh2021}. The test examples are then used in the experiments to obtain predictions, in which the influence of the attention scores is examined. For comparing attention score distributions and prediction vectors, we use two measurements. For quantifying the divergence between two attention distributions $\alpha_1$ and $\alpha_2$, we use the \emph{Jensen-Shannon-Divergence} (JSD)~\cite{Jain2019} that is based on the \emph{Kullback Leibler Divergence} (KL) defined in~\cite{kullback}: 

\begin{equation*}
    JSD(\alpha_1, \alpha_2) = \frac{1}{2}KL(\alpha_1 \parallel \frac{\alpha_1+\alpha_2}{2})+\frac{1}{2}KL(\alpha_2 \parallel \frac{\alpha_1+\alpha_2}{2}).
\end{equation*}

The change between two prediction vectors $p_1$ and $p_2$ is measured as the \emph{Total Variation Distance} (TVD):

\begin{equation*}
    TVD(p_1, p_2) = \frac{1}{2} \cdot \sum_{i = 1}^{\vert A\vert} \vert p_{1}^{i} - p_{2}^{i}\vert, \textnormal{where $p^{i}$ denotes the ith component of $p$.}
\end{equation*}

\phantomsection
\label{experiment1}
\paragraph{\textbf{Experiment 1 -- Attention Mechanism Parameter Manipulation:}}

\begin{figure}[t]
    \centering
    \includegraphics[scale=0.70]{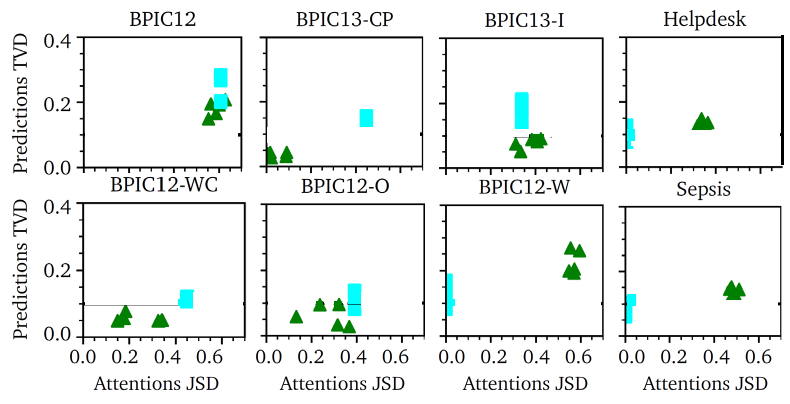}
    \caption{JSD vs. TVD plots. Rectangles = models with frozen weights, triangles = seeded base models. JSD can only take values between 0 and $log(2) = 0.693$.}
    \label{fig:attention-score-evaluation}
    
\end{figure}

We first investigate whether attention is necessary in the first place, following the experimental setup in~\cite{Wiegreffe2019}. We thus train a baseline model $\mathcal{M}_b$ with randomly initialized attention parameters (i.e., matrices $W_Q$, $W_K$, $W_V$, and $W_O$) and a modified one $\mathcal{M}_m$ identical to the training setup of $\mathcal{M}_b$ except that we freeze the attention parameters to uniform weights during training. We repeat this procedure five times with differently initialized baselines and modified models to minimize the influence of chance. To ensure to get indeed different initializations, we use random seeds. Then each manipulated transformer $\mathcal{M}_m$ is compared against the corresponding baseline model $\mathcal{M}_b$. To do this, we send all test samples through all models and obtain for each sample and model the corresponding attention distribution as well as the prediction vector. Then for each sample we compare attention distribution and prediction vector obtained from $\mathcal{M}_b$ with those obtained from each modified model using JSD and TVD, respectively. The computed JSD and TVD values are afterwards averaged across all samples. The intention behind this setup is, that the higher the divergence in attention scores (larger JSD) and the smaller the divergence in output (lower TVD), the less they affect the prediction~\cite{Wiegreffe2019,Jain2019}. Therefore, if the same prediction can be made with completely different attention scores, they can hardly serve as explanation. 

For interpretation, we plot the mean JSD against the mean TVD achieved for each model (see Fig.~\ref{fig:attention-score-evaluation}). Models positioned rightward and lower in the plots indicate less reliable attention scores. Hence, most of the event logs (BPIC12-W, Helpdesk, Sepsis) show good results. In the case of a right tendency (BPIC12), the TVD values are clearly greater than zero. Consequently, the attention scores can be trusted using the transformers trained on these event logs.



\phantomsection
\label{experiment2}
\paragraph{\textbf{Experiment 2 -- Attention Score Masking:}}

\begin{figure}[t]
    \centering
    \includegraphics[width=\textwidth]{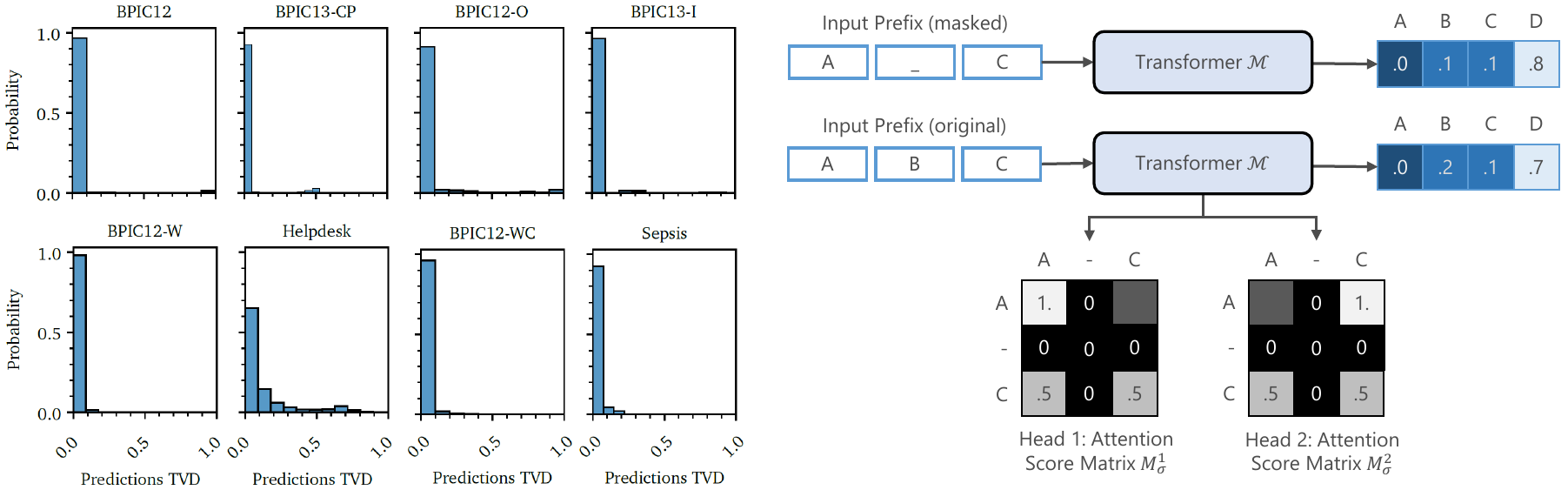}
    \caption{\textit{Left:} Variation between predictions TVD between masked elements in the prefix and only masked attention scores matrix. \textit{Right:} Used masking variants.}
  
    \label{fig:attention-score-masking-evaluation}
\end{figure}
In a second experiment, we investigate how particular attention values affect the prediction (see Fig.~\ref{fig:attention-score-masking-evaluation}). Therefore, we once mask elements in the input prefix and once the corresponding attention scores in the heads' attention score matrices and compare the model outputs via TVD. We mask elements in the prefix by replacing them by a padding symbol (\_) that indicates the model that there is no element. In contrast, masking the attention scores is done by setting all values of the corresponding rows and columns in the heads' attention score matrices to zero. 
Again we use all test samples and mask in each sample one element after the other. For each sample we obtain a prediction vector $p_m$ for the masked sample and a prediction vector $p_{am}$ for the unmasked sample that was processed with masked attention matrix. We compare $p_m$ and $p_{am}$ computing TVD. Low TVD values indicate that it does not matter whether we mask an element in the trace or in the attention score matrices. Hence, changes in the attention score matrices approximately affect the output in the same way as changes in the input trace. From this it can be concluded that particular attention values have relevance and can serve as an explanation. We observe that the histograms of TVD values (see Fig.~\ref{fig:attention-score-masking-evaluation}) are heavily left-leaning, i.e., masking directly in the prefix leads to almost the same prediction as masking in the attention score matrices. Minor deviations from this behavior are only noticeable in the histograms of the Helpdesk log. 



\vspace{-8pt}
\section{Explanation Approaches}
\label{sec:concept}

In this section we propose two global explainers that receive as input the trained transformer prediction model $\mathcal{M}$ and a set of prefixes $L$ to be explained. Both explainer construct a comprehensible directed graph describing the control flow of the process that acts as an indicator to which extent the PBPM model understands the control-flow of the process. 
Since attention scores reliably predict the next activity (cf. Section~\ref{sec:pre-study}), we can base our explanation approach on them.






\subsection{Backward Explainer}
The first explainer derives explanations for individual traces, subsequently integrating them into an overarching explanation for $L$ via a directed graph $G$. This process is explained in the remainder of this section and involves creating a local graph $G_{\sigma}$ for each prefix $\sigma \in L$, which is then merged directly into $G$. 
\begin{figure}[t]
    \centering
    \includegraphics[scale=0.5]{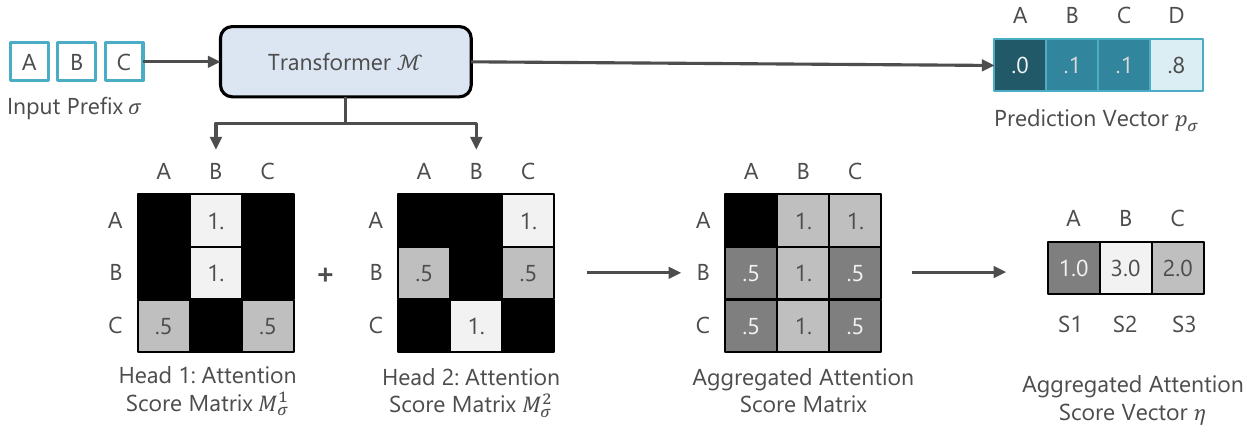}
    \caption{Determining aggregated attention scores for each event in the prefix, assuming that $\mathcal{M}$ possesses two heads.}

    \label{fig:relevant-activities}
\end{figure}
\paragraph{\textbf{Creating local graph:}} A crucial step for creating a local graph $G_{\sigma}$ for a prefix $\sigma$ is to use the head attention scores to identify the relevant activities $\mathcal{A}_r$ for the prediction $p_{\sigma}$. To do this we proceed as follows: We consider an activity as relevant for prediction $p_{\sigma}$, if its aggregated attention score over multiple modifications of $\sigma$ is large enough. To obtain different modifications of $\sigma$, we randomly apply modification operations. We pass each modification $\sigma_m$ to $\mathcal{M}$ to obtain prediction vectors $p_{\sigma_m}$ and the heads' attention score matrices $M_{\sigma_m}^{i}$. Then the cosine distance between $p_{\sigma_m}$ and $p_{\sigma}$ is computed and compared with a an a-priori defined threshold $\delta_{sim}$, to check whether the prediction for $\sigma_m$ is still close to the prediction of $\sigma$. If $\sigma_m$ satisfies this condition, the attention scores for each element $j$ in $\sigma_m$ are aggregated across the different heads:

\begin{equation*}
    S_j = \sum\limits_{n=1}^{\vert\sigma_m\vert}\left(\sum\limits_{i=1}^{n} M_{\sigma_m}^{i} \right)_{nj}.
\end{equation*}

In this formula first the heads' attention score matrices $M_{\sigma_m}^{i}$ are component-wise summed, and afterwards the $j$th column is summed. Hence, we get for each $\sigma_m$ an aggregated attention score vector $\eta_{\sigma_m} = (S_1, ..., S_{\vert \sigma_m \vert})$ containing a total attention score for each event. This procedure is illustrated in Fig.~\ref{fig:relevant-activities}. Next this total values are further aggregated per activity to a vector $\psi$ by summing up the total attention scores of each event belonging to a certain activity. Then $\psi_{\sigma}$ is normalized so that its components are within range $[0,1]$. We denote with $\psi_{\sigma}(a)$ the total attention score for activity $a$. Finally, we filter the activities by a given threshold $\delta_{attr}$ by keeping only activities with $\psi_{\sigma}(a) > \delta_{attr}$ to get only those activities that have a particularly high total attention score. This filtering is needed as irrelevant elements most have a small but not zero attention score and thus would be considered otherwise.
\begin{figure}[t]
    \centering
    \includegraphics[scale=0.5]{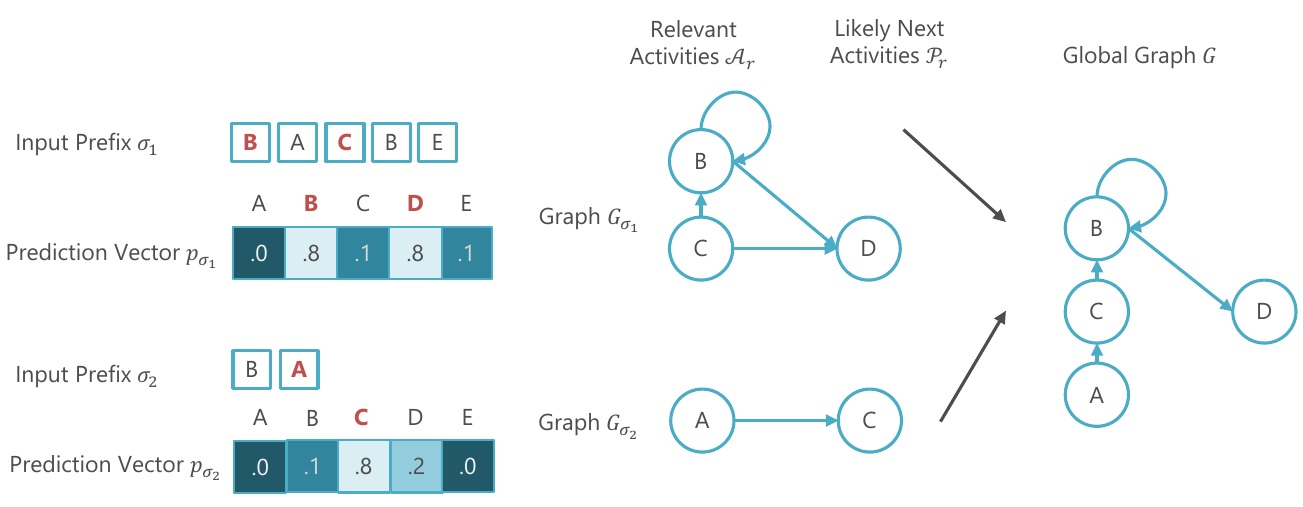}
    \caption{Example for BackwardExplainer. Relevant activities and likely next activities are highlighted bold faced in red color in the input prefix and prediction vector.}
  
    \label{fig:backward-explainer}
\end{figure}
The identified relevant activities $\mathcal{A}_r$ and the most likely next activities $\mathcal{P}_r$ obtained from prediction vector $p_{\sigma}$ (i.e., activities $a$ with $p_{\sigma}(a) > \delta_{pred}$). We do not use a fixed number of activities from $p_{\sigma}$, since depending on the process, in some cases the model is nearly sure that only one activity is possible, while in other cases (e.g., by an AND split) several activities seem valid to be executed next. Afterwards, edges between all nodes from $\mathcal{A}_r$ and $\mathcal{P}_r$ are established. Let us illustrate the procedure on the following example (see Fig.~\ref{fig:backward-explainer}). We have given prefix $\sigma_1 =\langle B, A, C, B, E \rangle$ where the aggregated attention scores identified activities $B$ and $C$ as relevant for the prediction. Moreover, we obtain a prediction vector $p_{\sigma_1}$ indicating that either activity $B$ or $C$ occur likely next. Hence, we get a local graph $G_{\sigma_1} = \left(\{B, C, D\}, \{(C, B), (B, B), (B, D), (C, D)\} \right)$. The graph can than be read as follows: For executing activity $D$ it is decisive that activities $B$ and $C$ were executed before. 
\paragraph{\textbf{Joining local graphs:}} Each local graph $G_{\sigma}$ is directly integrated into the global graph $G$. Therefore, vertices and edges of $G_{\sigma}$ are inserted into $G$ if not yet included. Assume that in our example in Fig.~\ref{fig:backward-explainer}, we have now a second local graph $G_{\sigma_2} = \left(\{A, C\}, \{(A, C)\} \right)$. Then the final graph would be $G = \left(\{A, B, C, D\}, \{(C, B), (B, B), (B, D), (C, D), (A,C)\} \right)$. After melting $G_{\sigma}$ into $G$, we perform a pruning step to remove undesired shortcuts that are likely not to occur in the underlying process. Such undesired shortcuts emerge, if an activity in the front part of $\sigma$ receive large attention for the prediction but activity and predicted activity do not follow each other directly. Hence, we remove an edge $(u, v)$ if edges $(u, a_n)$ and $(a_n, v)$ exist, with $a_n$ being the last activity in $\sigma$. 




\subsection{Attention Exploration Explainer}
The Backward Explainer has one major drawback: It directly determines the edges of the graph when analysing a prefix. Once inserted an edge is retained, except if it is a shortcut that can be removed in the pruning step. Thus, existing edges are not updated based on information obtained from further prefixes. We address this issue, by no longer relying on local graphs. Instead, we calculate a new relevance score for each activity across all prefixes presented to the algorithm. The decision about inserting an edge is postponed to the end of the procedure after inspecting all prefixes using the final relevance scores of the activities. 
\paragraph{\textbf{Determining relevance:}} To prevent relevance scores from increasing with each additional prefix and to reduce them when the attention score of an activity is consistently low, we introduce negative attention scores. In the transformer architecture, attention scores are inherently non-negative, with lower values indicating minimal impact. We encode the latter with a negative value. A positive value supports edge insertion, while a negative value is against it. 

Therefore we first determine for each prefix $\sigma\in L$ a so called \emph{score matrix} $K_\sigma$, with a column and a row for each activity $a \in A$. First, we extract for $\sigma$ the relevant activities $\mathcal{A}_r$ in the same way as in the Backward Explainer above. Additionally, we examine the positions of the relevant activities in $\sigma$ and store them in a set $\mathcal{I}_{\sigma}$. Next, we use all subsets of $\mathcal{I}_{\sigma}$ within two masking scenarios to quantify the individual and collective impact of activities on the prediction. For each subset $r \in \mathcal{I}_{\sigma}$, we consider two scenarios. In the "masking out most" scenario we mask the unimportant activities, i.e., all positions that are not contained in $r$. In contrast the "masking out a few" scenario masks the important activities, i.e., all positions in $r$. To obtain values for our score matrix $K_\sigma$, we now compare $\sigma$ with all of its masked versions (denoted as $\sigma_m$). Therefore, we send $\sigma$ and $\sigma_m$ through $\mathcal{M}$ to obtain prediction vectors $p_{\sigma}$ and $p_{\sigma_m}$ respectively. As well as the most likely next activities $\mathcal{P}_r$ for prefix $\sigma$. Then, we compute as in the Backward Explainer the per activity aggregated attention score vectors $\psi_{\sigma}$ and $\psi_{\sigma_m}$ indicating how much focus the model gives to an activity in the normal and masked trace, respectively. The computation of the scores is described in detail in Alg.~\ref{alg:relevanceScore}. In this calculation, a score is calculated for each activity in the masked prefix regardless of whether it is masked or not. For masked activities, the score is the product of the prediction score and attention scores. If the prediction for a masked activity matches its prediction in the normal trace, the score is negated, indicating reduced relevance due to the activity's negligible impact on the prediction. For non-masked activities, scores are derived differently: if an activity's prediction value is unchanged with masking, the score is the product of its attention score and normal trace prediction. If the prediction changes, the score is based on the product of the absolute differences in both attention scores and predictions, reflecting the influence of the activity in the masked trace. Eventually the calculated score is added to the corresponding cell (row for the predicted activity and column the masked activity) in the score matrix. 

\setlength{\textfloatsep}{2pt}
\begin{algorithm}[t!]
\caption{computeRelevanceScore}\label{alg:relevanceScore}
\KwData{Prefix $\sigma$, Masked prefix $\sigma_m$, aggregated attention scores per activity $\psi_{\sigma}$ and $\psi_{\sigma_m}$, prediction vectors $p_{\sigma}$ and $p_{\sigma_m}$, most likely next activities $\mathcal{P}_r$}
\KwResult{Score matrix}
$K \gets (0)_{ij} \forall i,j \in \{1,..., \vert A \vert\}$ \\
\ForEach{$a \in \mathcal{P}_r$}{
\ForEach{masked activity $a_m \in \sigma_m$}{ 
$s \gets p_{\sigma}(a) \cdot \psi_{\sigma}(a_m)$ \\
\If{$p_{\sigma}(a) \sim p_{\sigma_m}(a)$}{
$s \gets -s$
}
$K(a, a_m) \gets K_{\sigma}(a, a_m) + s $
}
\ForEach{non-masked activity $a_n \in \sigma_m$}{

\eIf{$p_{\sigma}(a) \sim p_{\sigma_m}(a)$}{$s \gets \psi_{\sigma_m}(a_n) \cdot p_{\sigma}(a)$}
{$s \gets \vert \psi_{\sigma}(a_n) - \psi_{\sigma_m}(a_n) \vert \cdot \vert p_{\sigma}(a) - p_{\sigma_m}(a)\vert $}
$K(a, a_m) \gets K_{\sigma}(a, a_m) + s $
}}
\Return $K$
\end{algorithm}
From Alg.~\ref{alg:relevanceScore} we receive a score matrix for each masked version of $\sigma$, which are summed to obtain two scenario-specific score matrices, $K_{\sigma}^{few}$ and $K_{\sigma}^{most}$. The two matrices comprehensively assess the relevance of each activity in $\sigma$, considering the impact of its presence or absence in different combinations.




\paragraph{\textbf{Building the explanation graph:}} Finally we must aggregate the score matrices of both scenarios we have obtained for each prefix: 

\begin{align*}
K^{few} &= \sum_{\sigma \in L} K_{\sigma}^{few} & K^{most} &= \sum_{\sigma \in L} K_{\sigma}^{most}.
\end{align*}
Afterwards we normalize each row in $K^{few}$ and $K^{most}$ (i.e., the entry in the row should sum up to one) to ensure comparability across different activities. Then, each entry of both matrices is converted to a boolean value, by checking whether it exceeds a predefined threshold $\delta_{edge}$ (true) or not (false). While "true" indicates a significant relevance of an activity on another, "false" indicates that there is no relationship between the activities. Eventually, we combine the insights from both masking scenarios by a component-wise logical OR operation between $K^{few}$ and $K^{most}$. The resulting matrix $K$ can be interpreted as adjacency matrix, where a "True" value indicates an directed edge from the corresponding activity of the column to the corresponding activity of the row. A "False" value indicates no edge. Finally, we construct the graph from this adjacency matrix.

\section{Evaluation}
\label{sec:evaluation}

In this section, we examine the performance of our explainers using the event logs from the pre-study. Our approach is implemented as prototypical Python framework that contains both explainers and all methods required for reproducibility\footnote{Source code: \url{https://github.com/mkaep/transformer-explainability}}. Since there are no other transformer-specific global explainers available, we cannot compare fairly with other approaches. A comparison with existing local or model-agnostic approaches would not be meaningful, as these pursue completely different objectives and global explainability is a much more difficult task. 
We first evaluate the quality of the models to rule out the possibility that they have not learned any skills at all. Only then, we evaluate the explainers, relying on quantitative metrics to avoid any subjective biases~\cite{Nauta2023}. This principle reduces the risk that our explainers will be blamed for the errors of the black box models.
\begin{table}[t]
\scriptsize{
	\begin{tabularx}{\textwidth}{l|X|X|X|X|X|X|X|X|X|X|}
		\toprule
  &
  \multicolumn{2}{X|}{Correctness} &   
  \multicolumn{2}{X|}{Completeness} & 
  \multicolumn{2}{X|}{Continuity}  &
  \multicolumn{2}{X|}{Contrastivity} &  
  \multicolumn{2}{X|}{Compactness} \\ \midrule
  
   & BW & AE & BW & AE & BW & AE & BW & AE & BW & AE\\
  \midrule 
  BPIC12      & 0.63 $\pm$0.20             & \textbf{0.65} $\pm$0.15  & 0.06 $\pm$0.01  & \textbf{0.07} $\pm$0.01 & 0.71 $\pm$0.05 & \textbf{0.93} $\pm$0.05 & \textbf{0.45} $\pm$0.33 & 0.26 $\pm$0.25 & \textbf{1.66} $\pm$0.20 & 4.12 $\pm$0.77\\
  \midrule
  BPIC12\_O     & \textbf{0.48} $\pm$0.22  & \textbf{0.48} $\pm$0.22 & 0.13 $\pm$0.02 & \textbf{0.21} $\pm$0.02 & \textbf{0.95} $\pm$0.07 & \textbf{0.95} $\pm$0.07 & \textbf{0.07} $\pm$0.18 & \textbf{0.07} $\pm$0.18 & 2.56 $\pm$0.06 & \textbf{2.21} $\pm$0.09\\
  \midrule
  BPIC12\_W &  \textbf{0.65} $\pm$0.21 & \textbf{0.65} $\pm$0.21 & 0.10 $\pm$0.08 & \textbf{0.27} $\pm$0.04 & 0.81 $\pm$0.13 & \textbf{0.92} $\pm$0.06 & \textbf{0.19} $\pm$0.30 & 0.16 $\pm$0.26 & \textbf{0.90} $\pm$0.14 & 1.86 $\pm$0.25\\
  \midrule
  BPIC12\_WC & \textbf{0.62} $\pm$0.18 & 0.56 $\pm$0.18 & \textbf{0.29} $\pm$0.10  & 0.28 $\pm$0.09 & \textbf{0.90} $\pm$0.06 & 0.89 $\pm$0.05 & \textbf{0.03} $\pm$0.13 & 0.02 $\pm$0.09 & \textbf{1.48} $\pm$0.60 & 1.89 $\pm$0.15\\
  \midrule
  BPIC13\_CP & \textbf{0.75} $\pm$0.18 & 0.60 $\pm$0.28 & 0.50 $\pm$0.14 & \textbf{0.79} $\pm$0.15 & 0.49 $\pm$0.13 & \textbf{0.98} $\pm$0.10 & \textbf{0.05} $\pm$0.20 & \textbf{0.05} $\pm$0.19 & \textbf{0.19} $\pm$0.24 & 2.09 $\pm$0.35\\
  \midrule
  BPIC13-I & N $\pm$ N & \textbf{0.57} $\pm$0.25 & 0.29 $\pm$0.04 & \textbf{0.77} $\pm$0.08 & 0.19 $\pm$0.24 & \textbf{0.99} $\pm$0.02 & \textbf{0.04} $\pm$0.18 & 0.03 $\pm$0.12 & \textbf{0.19} $\pm$0.24 & 2.00 $\pm$0.09\\
  \midrule
  Sepsis & \textbf{0.86} $\pm$0.13 & 0.59 $\pm$0.17 & 0.04 $\pm$0.02 & \textbf{0.15} $\pm$0.02 & 0.57 $\pm$0.09 & \textbf{0.92} $\pm$0.13 & \textbf{0.17} $\pm$0.33 & 0.13 $\pm$0.27 & \textbf{0.83} $\pm$0.16 & 2.76 $\pm$0.42\\
  \midrule
  Helpdesk & \textbf{0.91} $\pm$0.08  & 0.73 $\pm$0.18 & \textbf{0.36} $\pm$0.05  & 0.30 $\pm$0.04 & 0.73 $\pm$0.15 & \textbf{0.85} $\pm$0.19 & 0.08 $\pm$0.22 & \textbf{0.20} $\pm$0.30 & \textbf{1.99} $\pm$0.39 & 2.55 $\pm$0.62\\
  \midrule
AVG & 0.7 & 0.60 & 0.22 & 0.36
 & 0.67 & 0.93 & 0.14 & 0.12
 & 1.23 & 2.44 \\
 \midrule
AVG(AE-BW) & \multicolumn{2}{c|}{-0.01} &   
  \multicolumn{2}{c|}{0.134} & 
  \multicolumn{2}{c|}{0.26}  &
  \multicolumn{2}{c|}{-0.02} &  
  \multicolumn{2}{c|}{1.21} \\

  \bottomrule
\end{tabularx}}
	\caption{Results for Attention Exploration Explainer (AE) and Backward Explainer (BE); cells: mean (first), std. deviation (second); $N$ = no value could be computed, bold faced = best values; compactness: length of right-hand side.}

	\label{table:overall-evaluation}
\end{table}

We trained and evaluated the models following the setup in the pre-study. For evaluation we score a model by a commonly used weighted F1-Score~\cite{Bukhsh2021,Camargo2019,Pasquadibisceglie2019}. We achieved a mean weighted F1-score across all event logs of 0.74 (minimal value 0.46 and maximal value of 0.93). Hence, there are no no-skill models and it is likely that attention scores have been properly trained. Significantly higher values are not achievable as the transformer is limited to the activity event attribute. Consequently, many cases are undecidable and are therefore predicted purely statistically on the basis of frequency of occurrence. However, incorrect predictions are unproblematic for our explanation approaches, as they are intended to elucidate the rationale behind the model's decision. This is particularly, but not exclusively, useful for a more precise error analysis (Section~\ref{sec:conclusion-and-future-work}).


For our quantitative evaluation, we utilize various metrics from~\cite{Nauta2023}. These metrics rely on explanations in form of rule sets, because of which it is necessary to transform our graphs into rule sets. Therefore, each vertex $v$ leads to a rule reflecting its direct successors $S_v$. Hence, we obtain a set of rules of the form $R: v \rightarrow S_v$ for each $v$ in the graph's vertices $V$. Note, that all rules have exactly one activity on the left-side. The rule can be interpreted that right sides include all possible predictions, given the left-hand side. In the following, we give a brief overview of the intent of the used metrics and refer for more details to~\cite{Nauta2023}:

\noindent\textbf{Correctness:} This metric evaluates whether the given explanation is truthful~\cite{Nauta2023}. Therefore one input feature (in our case an activity) at a time is masked to determine its importance for model and explainer. Finally, the correlation between the feature importance for the model and for the explainer is calculated. Hence, the metric takes values within $[-1,1]$, where large values indicate that the explainer primarily uses the correct feature for its explanations. 

\noindent\textbf{Completeness:} This metric indicates how well explanations match the predictions of the transformer~\cite{Nauta2023}. 
Transformer predictions are considered ground truth and the right-hand sides of the explanation rules are evaluated as predictions. From the resulting (multi-label) confusion matrix we derive the F1-Score.

\noindent\textbf{Continuity:} This metric evaluates if the XAI approach provides similar explanations for slightly modified inputs and how well the approach generalizes w.r.t. unseen inputs~\cite{Nauta2023}. Varying the input it is checked how much the new explanation deviates~\cite{Nauta2023}. Values range between 0 and 1 (perfect continuous prediction)~\cite{Nauta2023}.

\noindent\textbf{Contrastivity:} Contrastivity represents the counterpart to continuity and evaluates how different the explanations of an XAI approach are for dissimilar inputs~\cite{Nauta2023}. Here, a value of 1 signals perfect contrastivity. 

\noindent\textbf{Compactness:} This metric evaluates the brevity of an explanation by counting the rules and averaging the lengths of their right-hand sides. Per~\cite{Nauta2023}, a more compact (shorter) explanation is preferable due to human cognitive limitations.

To apply the metrics, a consistent percentage of prefixes from each event log is randomly chosen to accommodate their varying sizes. Metrics are computed for each prefix individually, followed by averaging these values. Table~\ref{table:overall-evaluation} presents these average values together with their standard deviations. We observe consistently high correctness values, which clearly indicate that high attention scores actually correlate with the activities that are decisive for the prediction. Only the BPIC12\_O log falls slightly. A more heterogeneous picture emerges in case of the completeness metric where, depending on the event log, relatively poor values (BPIC12) or quite good values (BPIC13) are achieved. A deeper investigation revealed that the low values are caused by low recall values. Given the significantly higher precision values, the given explanations are often correct, however not all explanations are found. In the outputted graphs (or rules) it means that the edges are correct (high precision), but some edges are missing (low recall). Finally, the consistently very high continuity values suggest that the explainers are very robust, i.e., they also provide meaningful explanations for previously unseen data. However, the simultaneously very low contrastivity values suggest that the changes in the input have hardly any effect on the prediction. This is essentially caused by the fact, that the transformer only learns the most frequent execution paths due to its limitation to the event activity attribute, i.e., it reacts to strong changes in a prefix by providing a "standard" prediction. 
For a holistic comparison that indicates the performance of our explainers across the different datasets, we aggregated the results as follows. First, we calculate for each event log the difference between the value of the particular metric achieved by the Attention Exploration Explainer and the Backward Explainer. Afterwards we average all difference values to get a measure of the performance differences across the event logs (cf. last row of Table~\ref{table:overall-evaluation}). We see that across all considered event logs the Backward Explainer and the Attention Exploration Explainer perform nearly identical in terms of correctness and contrastivity. Nevertheless, there are some differences for particular event logs, e.g., in case of contrastivity for BPIC12 and Helpdesk) as well as in correctness for BPIC13\_CP, Sepsis and Helpdesk. This slightly poorer performance on these event logs for correctness metric is attributed to its higher recall values in the completeness metric. This results in the Attention Exploration Explainer identifying more explanations that are occasionally erroneous. This is particularly problematic in variant-rich processes (Sepsis), where explanations are often applicable to only a few traces, and event logs with concept drifts (Helpdesk), where explanations may lose their validity over time. For completeness and continuity, however, the Attention Explainer performs noticeably better. Notwithstanding the higher compactness values for the Attention Exploration Explainer, they remain within a low range ($<5$) and therefore do not impede interpretability. In contrast, the Backward Explainer’s low compactness values are one of the reasons for performing worse in completeness and continuity. As superior compactness of explanations is of little value in the absence of strong correctness, completeness, and continuity, it can be considered as a secondary metric. In conclusion, the Attention Exploration Explainer can be considered the superior option in the overall evaluation.

\section{Concluding Remarks}
\label{sec:conclusion-and-future-work}
This paper presents two transformer specific XAI approaches based on attention scores for next-activity prediction. We show first, that attention scores indeed form are a reliable basis for explanations and presented two approaches explaining predictions through graphs. These graphs show how attention scores for activities relate to predictions. 
The evaluation using various quantitative metrics demonstrates that the Attention Exploration Explainer outperforms the simpler Backward Explainer. This provides compelling evidence that explainers based on attention scores hold significant potential. This offers a wide range of useful opportunities for the future development and application of transformer-based predictions of the next activity. First, it provides transparency in the decision-making process of the model increasing the trust of process participants in the predictions. Second, the global explanatory capabilities of the presented approaches allow to assess the process understanding in general and to avoid contradictory explanations across different samples as in many local explainers. Third, the insights provided by the explainers, particularly in instances of erroneous predictions, offer a valuable starting point for model improvement, which may be achieved through enhanced training or the incorporation of additional training data. Moreover, an analysis of the graph structure can provide indications of potential weaknesses that may lead to the generation of erroneous predictions.

Therefore, we answer our research question (Section~\ref{sec:introduction}) as follows: the transformer's attention scores are effective for checking and visualizing its process understanding. However, since the trained prediction models only incorporate the activity event attribute, the process cannot be fully learned inherently.


\noindent\textbf{\emph{Limitations:}} Real-life event logs often suffer from quality issues such as noise, incompleteness, and inconsistencies, affecting the prediction and consequently the explanation quality in a negative way. 
In our pre-study we aggregate the attention scores of all heads to take a holistic perspective on the attention scores. Thus, it is unexplored what particular heads learn and how they potentially differ. In future work this should be investigated more in detail. Currently, the proposed explainers employ various thresholds to filter out less relevant relationships affecting the edges in the explanation graph. Thus, optimizing these thresholds holds vast potential to increase the explanation capabilities. An interesting starting point would be to adjust thresholds flexibly depending on the prefix considered. 

\noindent\textbf{\emph{Future Work:}} In addition to the points already mentioned for future work, the used masking and modification operations within the relevance determination should be improved by using more advanced operations like in the LORELEY approach. Also the generation of process models out of prediction models should be consequently fostered as a novel way of process discovery.

\bibliographystyle{splncs04}
\bibliography{literature}

\end{document}